\documentclass{article}

\usepackage[preprint,final]{neurips_2025}

\usepackage[utf8]{inputenc} 
\usepackage[T1]{fontenc}    
\usepackage{hyperref}       
\usepackage{url}            
\usepackage{booktabs}       
\usepackage{amsfonts}       
\usepackage{nicefrac}       
\usepackage{microtype}      
\usepackage{xcolor}         
\usepackage{natbib}
\usepackage{algorithm}
\usepackage{algorithmic}
\usepackage{wrapfig}
\usepackage{amsmath}
\usepackage{graphicx}
\usepackage{subfig}
\usepackage[most]{tcolorbox} 
\usepackage{listings}
\usepackage{hyperref}
\title{Automated Reward Design for Gran Turismo}

\author{%
  Michel Ma\thanks{Work done during an internship at Sony AI.} \\
  Mila, University of Montreal\\
  \texttt{michel.ma@mila.quebec}\\
  \And
  Takuma Seno\\
  Turing Inc.\\
  \texttt{takuma.seno@turing-motors.com}\\
  \And
  Kaushik Subramanian\\
  Sony AI\\
  \texttt{kaushik.subramanian@sony.com}\\
  \And
  Peter R.~Wurman\\
  Sony AI\\
  \texttt{peter.wurman@sony.com}\\
  \And
  Peter Stone\\
  Sony AI, UT Austin\\
  \texttt{peter.stone@sony.com}\\
  \And
  Craig Sherstan\\
  Sony AI\\
  \texttt{craig.sherstan@sony.com}\\
}

\begin{document}

\maketitle

\begin{abstract}
When designing reinforcement learning (RL) agents, a designer communicates the desired agent behavior through the definition of reward functions - numerical feedback given to the agent as reward or punishment for its actions. However, mapping desired behaviors to reward functions can be a difficult process, especially in complex environments such as autonomous racing.
In this paper, we demonstrate how current foundation models can effectively search over a space of reward functions to produce desirable RL agents for the Gran Turismo 7 racing game, given only text-based instructions.
Through a combination of LLM-based reward generation, VLM preference-based evaluation, and human feedback we demonstrate how our system can be used to produce racing agents competitive with GT Sophy, a champion-level RL racing agent, as well as generate novel behaviors, paving the way for practical automated reward design in real world applications.
\end{abstract}


\section{Introduction}
Significant algorithmic progress in reinforcement learning has led to impressive results in various domains such as robotics and video games \citep{tang2024rlrobots, badia2020agent57outperformingatarihuman, wurman2022outracing}. In particular, GT Sophy \citep{wurman2022outracing} used RL to train an agent from scratch to outrace top human competitors in the Gran Turismo video game, marking a significant achievement in applying RL to complex, high-fidelity simulated environments. 
Achievements like GT Sophy are built on algorithms that maximize a predefined scalar objective known as the reward function. Thus, the reward function is the primary tool for practitioners to use to communicate a desired behavior to an agent.
However, even for experts, reward function design remains a difficult process \citep{knox2023reward} and is a significant barrier to the wider adoption of RL.


When RL is applied to novel tasks, we must first define an appropriate reward function, a process called reward design \citep{singh2009rewards} or reward shaping \citep{ng1999policy}.
Reward design remains largely a manual process of trial and error requiring both RL expertise and specific domain knowledge \citep{booth2023perils,knox2023reward}. Often times, desired behaviors are easily communicated between humans through natural language and yet the same behaviors cannot be readily expressed through numerical functions.
For example, designing a reward function for a humanoid to ``\textit{perform a back flip}" or for an autonomous racing agent to ``\textit{win races while maintaining good sportsmanship}" is well understood and recognizable by most individuals, yet difficult to express quantitatively either due to the complexity of encoding the behavior or because of the inherent inexactness of the behavior.

Recently, large-scale foundation models have shown impressive capabilities in code generation, reasoning, image processing, and natural language understanding across a wide range of domains \citep{singh2022progprompt,wei2022chain,roziere2023code,awais2025foundation}. The increased accessibility and performance of these models has led to a growing body of work leveraging the near expert-level knowledge contained within large language models (LLMs) and video language models (VLMs) for reward design in reinforcement learning \citep{maeureka,xietext2reward,hazra2024revolve,klissarov2023motif,escontrela2023video,sontakke2023roboclip,li2024auto,carta2022eager}. One promising class of approaches to automate the conversion from text-specified behaviors to fully trained agents is to leverage LLMs to turn goals such as ``\textit{win races while maintaining good sportsmanship}" into appropriate reward functions as code amenable for any out-of-the-box RL algorithm to optimize.

Of particular interest is Eureka \citep{maeureka}, an LLM-based approach that uses an evolutionary algorithm to generate and train multiple reward functions in parallel; an approach which would otherwise be very labor intensive for any one human to perform. However, their method presupposes access to a ground-truth \textit{fitness metric} that provides a numerical signal for assessing the success of a given policy. Not only can it be computationally expensive to train agents for each generated reward function, but fitness metrics are not always available, and designing this metric is not necessarily simpler than directly designing the reward function itself. Alternatively, one could rely exclusively on human feedback as a replacement for the fitness metric \citep{xietext2reward,hazra2024revolve}. Unfortunately, human feedback inherently requires a certain amount of manual effort that scales unfavorably with the number of reward functions generated, largely defeating the purpose of an automated search system. 

In this paper, we set out to develop a general purpose LLM-based system that can efficiently and reliably produce RL-trained agents for Gran Turismo given only text-based inputs from a user.
To this end, we introduce a scalable iterative framework for automated reward design without the need for a \textit{fitness metric}. We hypothesize that current foundation models can accurately distinguish good from bad behavior just as well as humans can. 
Therefore, after each iteration, we elicit preferences from a VLM over a population of learned policies to select the best-performing reward function, serving as a replacement to the \textit{fitness metric}. The resulting set of preferences is used to automatically prune misaligned candidate reward functions for future iterations via a trajectory alignment coefficient \citep{muslimani2025towards}.

We use GT Sophy, a superhuman agent trained with expert-designed reward functions, as a useful point of reference in the Gran Turismo 7 environment for our experiments. GT 7 is a video game built around a complex racing simulator grounded in reality for which it has historically been difficult to design appropriate reward functions \citep{wurman2022outracing}. 
We show that our framework can replace countless hours of expert reward design, generating reward functions capable of producing a racing agent similar in performance to current state-of-the-art agents with minimal human intervention. We also conduct experiments to better understand the contributions of the VLM's preferences towards the final results, and examine the novelty of the generated reward functions. Finally, our framework is used end-to-end to produce novel behaviors in GT 7 directly from text, eliminating the need for manual reward design.

\vspace{-0.5em}
\section{Related Works}
\vspace{-0.5em}
\paragraph{RL for Autonomous Racing.} In the first instances of using RL for autonomous racing \citep{fuchs2021super,remonda2021formula}, the agent's goal, to minimize lap times, was straight-forward to implement for RL, because no other car were on the track. However, as progress was made towards developing agents in a multi-car race setting \citep{song2021autonomous,wurman2022outracing,lee2025champion}, reward functions became more complex to account for the subtleties of motor-sport racing rules \citep{macglashan2022sportsmanship}. These challenges are not unlike those in the field of autonomous driving \citep{chen2024end}, where additional considerations must be dealt with to ensure reasonable behaviors within the broader context of other humans \citep{knox2023reward,hazra2024revolve}. To the best of our knowledge, our work is the first of its kind to push the edge of superhuman performance in an autonomous racing environment with an automated reward design system.

\vspace{-0.5em}
\paragraph{Coding Reward Functions with LLMs.} A number of existing works have used LLMs to code reward functions for RL agents \citep{maeureka,li2024auto,zhang2024orso,sun2024large,yu2023language,xietext2reward,hazra2024revolve,han2024autoreward} in robotics tasks \citep{maeureka,sun2024large,yu2023language,xietext2reward}, video games \citep{li2024auto} and autonomous driving \citep{han2024autoreward,hazra2024revolve}. The reward generation process in these works largely follow the same blueprint: the environment and task description are fed as text-based context to the LLM, and it must then produce reward code either from scratch or following an existing template to be used for an  RL algorithm. Where they differ is mostly in their applications and their automated evaluation process. Notably, several of these works \citep{maeureka,sun2024large,yu2023language,li2024auto,zhang2024orso} assume access to some oracle sparse reward function or fitness metric to automatically provide feedback about a reward function's performance. Others \citep{xietext2reward,hazra2024revolve,han2024autoreward} without such access rely heavily on a human's feedback for each generated agent for evaluation. In contrast, our work attempts to generate superhuman agents with automated LLM reward design in Gran Turismo without access to a fitness metric or oracle reward, while limiting the amount of human feedback necessary.
\vspace{-0.5em}
\paragraph{LLMs and VLMs as the Reward Function.} Alternatively, others have suggested using LLMs \citep{klissarov2023motif,klissarov2024maestromotif,kwon2023reward} and VLMs \citep{escontrela2023video,huang2024vlm,yang2024adapt2reward,sontakke2023roboclip,rocamonde2023vision} more directly either as a reward function, or as a supplementary shaping reward to the existing rewards of the environment instead of going through the intermediate step of coding a reward function. In all these  cases, some reward function is extracted from the foundation models through preference elicitation \citep{klissarov2023motif,klissarov2024maestromotif,kwon2023reward} or by using the embedding space of the foundation model to numerically quantify behaviors \citep{escontrela2023video,huang2024vlm,yang2024adapt2reward,sontakke2023roboclip,rocamonde2023vision}. In either case, such methods do not generate interpretable reward functions, and necessitate a white-box foundation model in the latter case in order to access a model's embedding space. While a part of our system elicits preferences from a foundation model similarly to \citep{klissarov2023motif} and uses a VLM to identify behaviors \citep{sontakke2023roboclip,huang2024vlm}, our method uses these components as a supporting module for a larger reward coding framework. 

\vspace{-0.5em}
\section{Problem Setting}
\vspace{-0.5em}
\label{sec:it-llm-rd}
\begin{wrapfigure}{R}{0.5\textwidth}
\begin{minipage}{0.50\textwidth}
\centering
    \includegraphics[width=1.\linewidth]{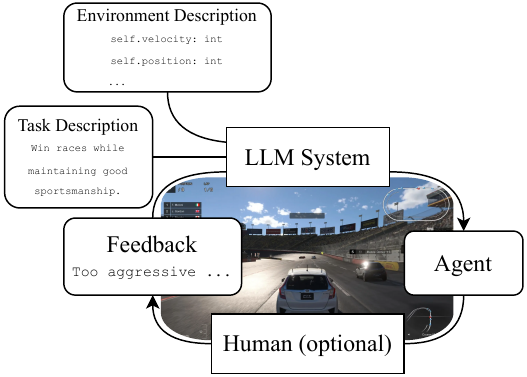}
    \vspace{-1em}
    \caption{\footnotesize Overview of an iterative LLM-based reward design pipeline. The environment and task description are fed into a black-box system to produce compliant agents, and the agent is fed back into the system for improvement, sometimes accompanied by some human feedback. }
    \label{fig:overall-iterations}
    \vspace{-1.5em}
\end{minipage}
\end{wrapfigure}

\paragraph{The Reward Generation Problem.} Defined as an extension of the \textit{reward design problem} \citep{singh2009rewards}, the \textit{reward generation problem} (RGP) \citep{maeureka} can be expressed by the tuple $P = \langle M, \mathcal{R}, \pi, F, g \rangle$, where $M = (S, A, T)$ contains the state space $S$, action space $A$ and transition function  $T$ of the environment. $\mathcal{R}$ is the space of all reward functions, $\pi$ is a policy optimizing a reward function $R \in \mathcal{R}$ in the resulting Markov Decision Process $(M, R)$, and $F: \Pi \rightarrow \mathbb{R}$ is a fitness function that quantifies the performance of a policy.  Suppose we have access to an algorithm $\mathcal{A}: \mathcal{R} \rightarrow \Pi$ to find a policy that optimizes any given reward function. The objective of this problem is to find a reward function $R$ such that $\mathcal{A}(R)=\pi^*$ maximizes $F$. The defining feature of an RGP that separates it from the typical reward design problem is the text-based nature of the problem. Specifically, the task description, or the agent's goal, is given as the string $g$, and the components $(S, A, T)$ are assumed to be fully expressible as code.

In this paper, we make the assumption that the fitness function $F$ is not accessible while searching for a reward function. In practice, this can be due to a multitude of reasons. Designing a fitness metric for a task can pose similar challenges to designing reward functions, where the definition of \textit{success} is either too vague or too complex to clearly define numerically. Other times, calls to the fitness metric can simply be too expensive to use during training, such as the case where $F$ is based on human expert evaluations. 



\paragraph{Iterative LLM-based Reward Design.}
One approach to addressing the \textit{reward generation problem} is to use an LLM to iteratively generate reward function code towards maximizing the fitness metric. Such an approach, which we name \textit{iterative LLM-based Reward Design}, follows the general structure in Figure \ref{fig:overall-iterations}. First, the environment and task descriptions are fed into some LLM-powered system that outputs a trained agent. Then, feedback is optionally collected from a human in the form of text, and both the agent and feedback are fed back into the main LLM system for improvement on the following iteration. This general framework encompasses several prior works \citep{maeureka,xietext2reward,hazra2024revolve,li2024auto,sun2024large,han2024autoreward} in LLM-based reward design, where the core systems for reward generation, evaluation and training can vary from method to method, but all follow an iterative principle to automatic reward design. As an additional benefit to these systems, the iterative process of designing reward functions can be useful for human users as well. In the cases where $F$ is held implicitly in the mind of a user, $g$ is often imperfect due to the vagueness of language, and it is only by observing agent behaviors that a user can iteratively refine its own expectation of an optimal agent.

\vspace{-0.5em}
\section{Method}
\vspace{-0.5em}

\begin{figure}[t]
    \centering
    \includegraphics[width=1.\linewidth]{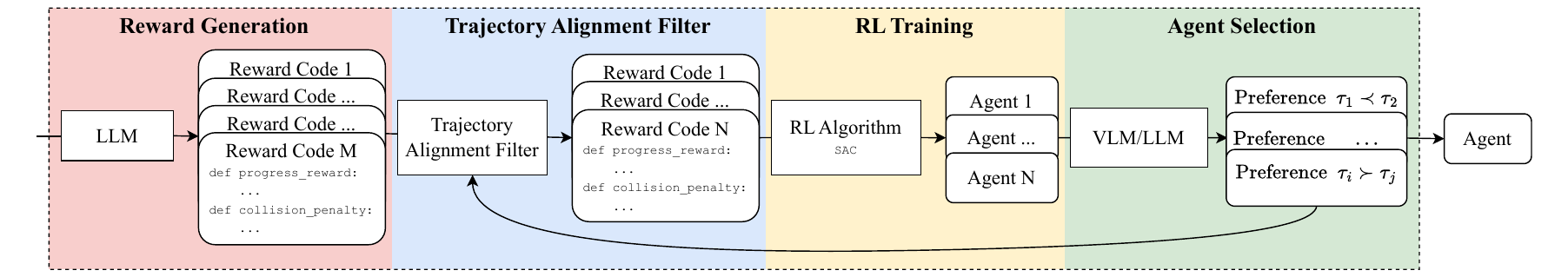}
    \caption{\footnotesize Expansion of the \textit{system} block in Figure \ref{fig:overall-iterations} for our framework. First, reward code is generated from scratch by an LLM. The rewards are then filtered through a trajectory alignment filter to avoid training misaligned rewards. The rewards are all trained separately by an RL algorithm. Finally, a VLM or LLM selects the best agent for further iterations. }
    \label{fig:ours-overview}
    \vspace{-1.5em}
\end{figure}

Our automated reward design system follows the same outer loop for \textit{iterative LLM-based Reward Design} described in Section \ref{sec:it-llm-rd} and Figure \ref{fig:overall-iterations}. For the environment description, other methods \citep{maeureka,xietext2reward,yu2023language,sun2024large} provide the full code that describes the environment as part of the LLM prompt. We argue that having environment code neatly contained as a single file or class is more an exception than the rule. Many complex environments, even those that are simulations, such as GT, can rarely be captured so concisely. To this end, we provide only the Python dataclass API of the state and action space of the Gran Turismo 7 training environment, which contains variable names, comments, and data types for each field.\footnote{We acknowledge that effort was put into ensuring that these objects were properly named and commented before being given to the LLM. }
Instead of providing code for the dynamics, our system relies on the LLM's internal world model of physics and racing to reason about the dynamics of our environment.
For the remainder of this section, we focus on the inner workings of our LLM system shown in Figure \ref{fig:ours-overview}. Subscripts are used to denote anything related to the $i$th generated reward function. 

\paragraph{Reward Generation.}

For the first iteration, the environment context and task description $g$ are combined into a single prompt that tasks an LLM to design reward functions for an RL algorithm. For future iterations, the prompt also contains information about the previous iteration's best agent (see \textbf{Agent Selection}): text-based feedback about the agent's behavior, the reward function $R_{best}$, and training diagnostics $\eta_{best}$ expressed as the reward function's value at different points during training. We ask the LLM to separate the reward function into multiple simpler components that are then combined by a weighted sum to obtain the final reward. The LLM is also tasked with setting the weights for each component. Reward functions are validated as \textit{"runnable"} by calling them on randomly sampled states and actions. Generated reward functions that fail this test are iteratively re-generated by the LLM with the error trace included in the prompt until the generated functions pass the test \citep{olausson2023demystifying,le2022coderl}. $M$ different reward functions are independently sampled during this step.

\paragraph{Trajectory Alignment Filter.}
Training RL agents can be quite expensive, especially in a complex environment such as Gran Turismo. To reduce the number of RL agents trained at every iteration, we use the trajectory alignment coefficient (TAC) \citep{muslimani2025towards} based on a Kendall's Tau-a correlation \citep{kendall1945treatment} to filter out potentially misaligned reward functions. To do so, assume access to a dataset of previously collected preferences, $(p_{ij}, \tau_i, \tau_j) \sim \mathcal{D}_{\text{pref}}$, where $\tau$ is a sub-trajectory of length $L$ of states and actions $(s_1a_1,s_2a_2, ..., s_L a_L)$, and $p_{ij} \in \{0, 1\}$ indicates a preference of $\tau_i \succ \tau_j$ if $p_{ij}=0$ and $\tau_i \prec \tau_j$ otherwise. Let $G(\tau, R_m)=1/L\sum_{t=1}^L R_m(s_t,a_t)$ be the average reward \footnote{We use an average reward for the TAC to account for varying lengths of sub-trajectories and due to the very long horizon nature of the task which resembles a continual task in practice.} over a sub-trajectory $\tau$ for any given reward function $R_m$ and $\tilde{p}_{ij}(R_m) \in \{0, 1\}$ is the preference induced by comparing $G({\tau_i}, R_m)$ and $G({\tau_j}, R_m)$. The trajectory alignment coefficient for a reward function $R_m$ is then calculated as:
\begin{align*}
    \sigma(R_m, \mathcal{D}_\text{pref})&= \frac{2P - |\mathcal{D}_\text{pref}|}{|\mathcal{D}_\text{pref}|} \enspace \text{where} \enspace 
    P:= \sum_{(p_{ij}, \tau_i, \tau_j) \in \mathcal{D}_\text{pref}} p_{ij}\tilde{p}_{ij}(R_m) + (1 - p_{ij})(1-\tilde{p}_{ij}(R_m)) \enspace .
\end{align*}
The TAC ranges from $[-1,1]$. Intuitively, when $\sigma_m=-1$, the reward function $R_m$ disagrees with all preference labels in $\mathcal{D}_\text{pref}$; when $\sigma_m=1$,  the preferences induced by $R_m$ are in total alignment; when $\sigma_m=0$, the reward function is no better than one that generates random values. Before training, only the top $N$ best aligned reward functions, as indicated by $\sigma_m$, are kept for the following training step, where $N$ is a hyperparameter smaller than the number of rewards sampled in the previous step $M$. In doing so, we hope to reduce computational costs associated with training, while searching over an equally large number of reward functions.

\begin{wrapfigure}{R}{0.52\textwidth}
\begin{minipage}{0.52\textwidth}
\vspace{-1.5em}
    \begin{algorithm}[H]
    \caption{Automated Reward Design for GT}
    \label{alg:system-alg}
    \textbf{Input: } Empty dataset of preferences $\mathcal{D}_{\text{pref}}$, RL algorithm $\mathcal{A}_E$, initial task prompt $\texttt{prompt}$.

    \begin{algorithmic}[1]

    \FOR{$i = 1$ to $K$}
        \STATE \textcolor{blue}{\texttt{// Generate M reward functions}}
        \STATE $R_1, R_2, \cdots, R_M  \sim  \texttt{LLM}(\texttt{\small prompt})$ 
    
        \STATE \textcolor{blue}{\texttt{// Compute the alignment coefficient for each reward}}
        \STATE $\sigma_m \leftarrow \texttt{Alignment}(R_m, \mathcal{D}_{\text{pref}})$
        \STATE \textcolor{blue}{\texttt{// Keep only the top N rewards}}
        \STATE $R_1, R_2, \cdots, R_N \leftarrow \texttt{top\_n}(R_m, \sigma_m )$
        \STATE \textcolor{blue}{\texttt{// Train an agent for each reward, and get a policy, trajectory, and training diagnostics}}
        \STATE $\pi_n^*, \tau_n, \eta_n  \leftarrow \mathcal{A}(R_n)$
        \STATE \textcolor{blue}{\texttt{// Get preference labels over the newly trained policies}}
        \STATE $p_{ij} \leftarrow \texttt{VLM}(\tau_{i}, \tau_j) \enspace \forall i, j \in \{N\}$ 
        \STATE \textcolor{blue}{\texttt{// Get Bradley-Terry scores from VLM labels}}
        \STATE $b_{1:N} \leftarrow \texttt{bradley}(\{p_{ij} \enspace | \enspace  \forall i, j \in \{N\})$
        \STATE \textcolor{blue}{\texttt{// Select best performing agent}}
        \STATE $best \leftarrow \text{argmax} \enspace b_{1:N}$
        \STATE \textcolor{blue}{\texttt{// Get human feedback for best policy}}
        \STATE $\texttt{\small feedback} \leftarrow \texttt{human}(\pi_{best})$
        \STATE \textcolor{blue}{\texttt{// Construct new prompt}}
        \STATE $\texttt{\small prompt} \leftarrow \texttt{\small prompt}  : \texttt{\small feedback} : R_{best}: \eta_{best}$
        \STATE \textcolor{blue}{\texttt{// Add new preferences to dataset}}
        \STATE $\mathcal{D}_{\text{pref}} \leftarrow \mathcal{D}_{\text{pref}} + \{p_{ij} \enspace | \enspace  \forall i, j \in \{N\}\}$
    \ENDFOR
    \RETURN $R_{best}$
    \end{algorithmic}
    \end{algorithm}
    \vspace{-3em}
\end{minipage}
\end{wrapfigure}

\paragraph{Training}
Any black-box RL algorithm can be used in this step so long as the following three outputs are obtained: a trained policy $\pi^*$, a sample trajectory of the trained policy $\tau$, and training diagnostics $\eta$. The diagnostics are represented as a string capturing the reward function's value at different points during training. For the first iteration, each of the $N$ reward functions are trained from scratch. In subsequent iterations, all models are trained from scratch again, except the replay buffer from the previous iteration's best agent (see \textbf{Agent Selection}) is used as a secondary replay buffer to speed up the learning process \citep{tirumalareplay}. The additional experience from the secondary replay buffer can help reduce reinforcement learning training times \citep{tirumalareplay,nikishin2022primacy,d2022sample}, which can pose significant computational challenges in an iterative approach where RL is part of the inner loop. An ablation is done on the impact of the secondary seed buffer in Appendix \ref{sec:buffer-ablation}.

\paragraph{Agent Selection}
After training each of the $N$ reward functions, our system is tasked to select the best performing policy for human evaluation. We elect to use a pairwise preference-based method due to its robustness \citep{ouyang2022training,wirth2017survey}. For each trained policy, a sufficiently long trajectory $\tau$ is sampled along with its corresponding video footage at ten frames per second. Then, a VLM is prompted to give a preference $p \in \{0, 1\}$ over all $\binom{N}{2}$ pairs of agents in the current iteration given both trajectory and video data for each agent (Appendix \ref{sec:vlm-prompts}). A Bradley-Terry model \citep{bradley1952rank} is used to rank the agents, and the best policy is kept for the next iteration. Finally, all preferences $p_{ij}$ and corresponding trajectories $\tau_i, \tau_j$ are saved in the dataset $\mathcal{D}_\text{pref}$. The complete pipeline of our framework is shown in Algorithm \ref{alg:system-alg}.

\section{Experiments}
\vspace{-0.5em}
All experiments are done in the Gran Turismo 7 video game racing environment in a single car-track scenario: Lake Maggiore - Full Course\footnote{Lake Maggiore is a popular track in GT 7 that features a diversity of interesting racing scenarios such as tight-harpins, both high and low speed corners, and a straight long enough to take advantage of slip stream.} with a Mercedes-AMG GT3 '16. For all code generation, we use the latest available reasoning model available to us at the time provided by OpenAI, \texttt{gpt-4o}. Similarly, all preferences collected during agent selection are done using the latest vision-enabled model provided by OpenAI, \texttt{gpt-4.1}. All human feedback is collected from one of the authors. 

Training of the RL agent uses the approach described in \citep{wurman2022outracing} -- a modified version of the soft actor-critic algorithm \citep{haarnoja2018soft}, QR-SAC.
On average, a production-level baseline GT Sophy agent is trained for over 1500 epochs\footnote{One epoch is equivalent to 6000 training steps.} from scratch, which takes approximately ten days to train on a single A100 GPU and a dozen Playstation 4 systems. For our experiments, we reduced the training time of each iteration to 300 epochs, and only train the policy from the final iteration for 1500 epochs without a secondary replay buffer (to compare with GT Sophy) for evaluation. From now on, we refer to policies trained for 300 epochs as \textit{intermediate} policies or agents, and  final policies trained for 1500 epochs without a secondary replay buffer as \textit{final} policies or agents. Three seeds of the final policy are trained for all experiments.
Note that intermediate policies are unlikely to have converged even with the use of a secondary replay buffer, but are usually sufficient for preference elicitation in the agent selection step. 

\begin{figure}[t]
    \centering
    
    \includegraphics[width=0.82\linewidth]{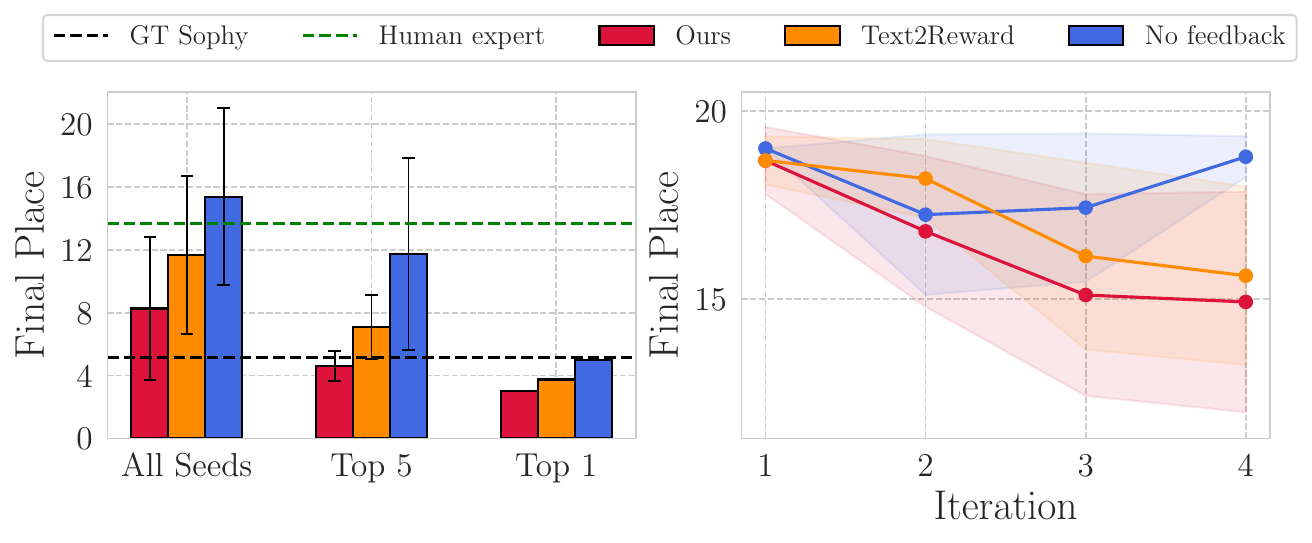}
        \includegraphics[width=0.17\linewidth]{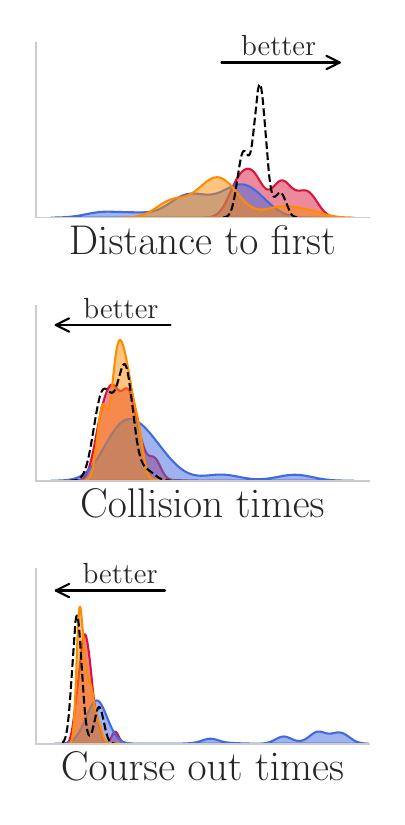}

    \vspace{-1em}
    \caption{\footnotesize \textbf{Left}: Average final placements ($\pm \text{std}$) of final policies over all ten seeds, top five seeds only, and the best seed respectively. \textbf{Middle:} Average final placements ($\pm \text{std}$) of intermediate policies over all reward functions trained for each iteration of every seed. \textbf{Right:} Distribution of various racing metrics evaluated with the final agents from the top five seeds of each method. }
    \vspace{-1em}
    \label{fig:main_results}
\end{figure}

\subsection{Outracing Human Experts}
\label{sec:main-exp}
The goal of our main experiment is to produce an agent competitive with GT Sophy: a superhuman agent that complies with all standard motorsport racing rules. Our GT Sophy baseline was trained using the current state-of-the-art reward function meticulously designed over time with the help of both RL experts and game designers to achieve superhuman performance. The reward function for this agent is detailed in Appendix \ref{sec:gt-sophy-rewards}. We also compare our agents to the performance of a human expert with over 25 years of experience in the Gran Turismo series who regularly ranks in the top 5\% of players in the world. 
For the RGP, we purposefully choose a vague task description $g$ to limit any prior bias we may have about what constitutes a good reward function in this task: "\textit{win races while obeying standard motorsport racing rules and maintaining good sportsmanship}".

We run three different variations of our framework for four iterations ($K=4$): (i) \textbf{Ours} generates ten different rewards ($N=10$) and trains only the top five rewards ($M=5$) from the trajectory alignment filter per iteration; \textbf{Text2Reward} generates only a single reward ($N=1, M=1$) per iteration, therefore skipping both the trajectory alignment filter and the agent selection steps which essentially recovers the Text2Reward framework \citep{xietext2reward}; (iii) \textbf{No feedback} also generates and trains ten and five reward functions respectively ($N=10, M=5$) per iteration, but instead replaces the human feedback by an LLM's feedback of the agent's behavior (Appendix \ref{sec:closed-loop-system}), making it a closed-loop system. All other hyperparameters follow the same parameters used to train GT Sophy.

\subsubsection{Evaluation}
For the same reasons that make reward design difficult for this task, evaluating agents for \textit{good sportsmanship} or \textit{rule abiding behavior} can be difficult, and is typically done using human expert judges for production-level agents. In order to systemically evaluate our agents in this work, we rely on 
two different racing statistics to provide a proxy for good sportsmanship and rule abiding behavior: \textit{car collision time}, the total amount of time spent in contact with other cars, and \textit{course out time}, the total amount of time spent off the legal racing area. Other works in autonomous racing also use collision times to assess proper sportsmanship \cite{lee2025champion,fuchs2021super,song2021autonomous}. We found it necessary to also consider the time spent off course, since some agents were observed to gain an unfair advantage by taking illegal shortcuts during the race.

Policies are evaluated by starting a 4-lap race against 19 built-in AIs, an agent deployed by the game developers, with the agent to be evaluated beginning the race in last place. A hundred different races are collected to account for the variability of behaviors in the built-in AIs, and only trajectories in the 50\% interquantile range of performance are kept. 
For evaluation, an agent's final place defaults to last place if it exceeds a predetermined threshold for \textit{car collision time} and \textit{course out time}. Following the procedure in \cite{lee2025champion}, these thresholds were chosen by training five seeds of a GT Sophy agent and selecting the highest \textit{car collision time} and \textit{course out time} among the five agents. In other words, agents who do not achieve a similar or better \textit{car collision time} and \textit{course out time} than GT Sophy are automatically disqualified and set to last place. By definition, all GT Sophy agents pass this threshold, therefore biasing our evaluations in favor of GT Sophy.
All placements follow a zero-based indexing scheme where the best final placement is 0, and the worst is 19.

\subsubsection{Results}
The results of our experiments shown in Figure \ref{fig:main_results} and Table \ref{tab:main-exp-table} tell us that LLMs can in fact produce agents which are not only competitive with GT Sophy, but can sometimes surpass the latter's performance. While our system does not achieve this performance every time, the added ability to automatically search over multiple reward functions per iteration clearly improves both the consistency and the ceiling of our method compared to the Text2Reward baseline, without incurring any additional manual labor. 

Interestingly, while the closed-loop system which relies solely on intrinsic feedback from the LLM is capable of producing expert-level reward functions, its average performance is markedly worse than a system incorporating human feedback; most likely due to its tendency to exceed the required thresholds for collision times and time spent off course. The progression of the generated reward functions' performances shown in the middle of Figure \ref{fig:main_results} provides some additional insight: reward functions tend to diverge in performance without human grounding.  

\begin{table}[!hbt]
\resizebox{\textwidth}{!}{
    \centering
    \begin{tabular}{c|c|c|c|c|c}
    \hline
     & Ours & Text2Reward & No feedback & GT Sophy& Human Expert\\
     \hline \hline
        Valid agents &  \textbf{10/10} & 8/10 & 3/10 & -- & --\\
    Better than Sophy & \textbf{3/10} & 1/10 & 1/10 & -- & --\\
        Best dist. to first & $\textbf{243.0} \pm 9.1$ & $269.1 \pm 24.3$ & $ 320.2 \pm 3.6$& $317.8 \pm 17.2$ &$548.5 \pm 50.3$\\
        \hline
    \end{tabular}
}
\vspace{0.5em}

    \caption{\small \textit{Valid agents} counts the number of times the final policy for each seed of our framework passes the thresholds for good sportsmanship and rule-abiding behavior. \textit{Better than Sophy} counts the number of agents where the average final placement during evaluation is \textit{higher} than GT Sophy. \textit{Best dist. to first} reports the \textbf{best} final agent's (out of 10 seeds)  average distance ($\pm$ std) to the first car in meters during evaluation.}
    \label{tab:main-exp-table}

    \vspace{-2em}
\end{table}

\subsection{Replacing Human Expertise with VLMs}

\begin{wrapfigure}{R}{0.5\textwidth}
\begin{minipage}{0.5\textwidth}
\vspace{-1em}
    \centering
    \includegraphics[width=0.99\linewidth]{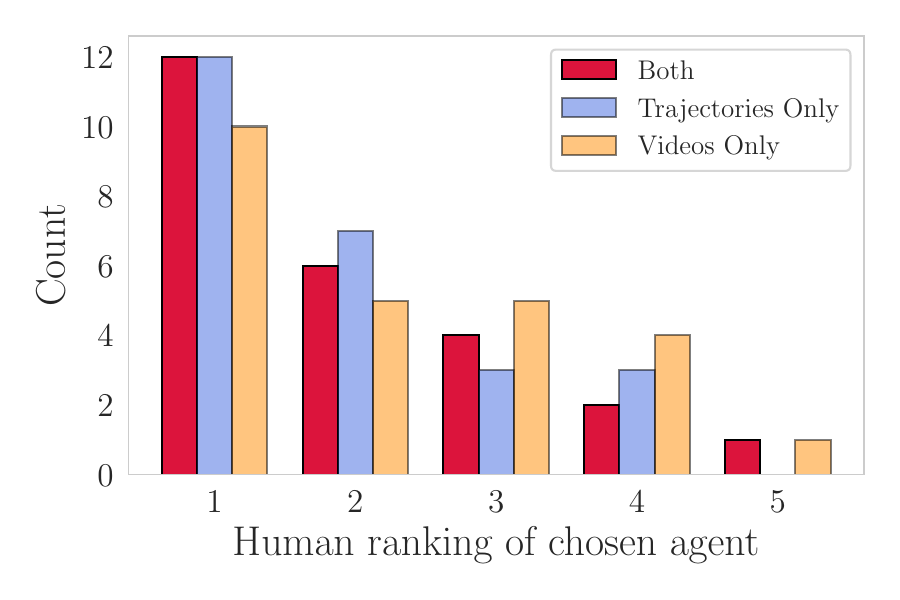}
    \vspace{-2em}
    \caption{\footnotesize \textbf{The VLM tends to select agents that are also favored by humans.} Frequency of a VLM's preferences, using different forms of inputs, to select the best, second best, third best, fourth best, and fifth best agent in a set of five agents, as decided by human experts.}
    \label{fig:vlm-prefs}
    \vspace{-1em}
\end{minipage}
\end{wrapfigure}
The preferences generated by the VLM during agent selection plays a crucial role in our framework. Aside from the direct impact they have towards replacing the fitness metric, the trajectory alignment filter also relies on its accuracy to prune reward functions. In the worst case, these preferences may be no better than random guessing, and the system would effectively be equivalent to generating a single reward function ($N=1, M=1$) per iteration.

In order to evaluate the accuracy of the preferences collected by the VLM, we first randomly select 10 iterations from those generated by "\textbf{Ours}" in the previous section. Each iteration contains five intermediate policies ($M=5$) and five corresponding trajectories from which we can create ten possible pairs of trajectories. Then, ten human subjects familiar with the Gran Turismo 7 environment were asked to blindly label a subset (twenty to thirty) of these pairs given the same task description $g$ defined above to create a small dataset of ground truth pairwise preferences, ensuring that each pair is labeled by at least two different individuals. 

To evaluate the VLM's image processing capabilities, we collect preferences from three different input modalities: (i) \textbf{VLM} collects preferences using both trajectory data and video data, which is the version used for the experiments in Section \ref{sec:main-exp}; (ii) \textbf{VLM-$\tau$} uses trajectory data only (no images); (iii) \textbf{VLM-$I$} collects preferences using only video data (images only). From the agreement rates shown in Table \ref{tab:vlm-prefs}, we observe that the preferences generated by the VLM have an agreement rate nearly as high as the percentage of times humans agree with each other, the latter of which can be seen as an upper bound for the VLM. The accuracy of the VLM also appears to drop significantly when given only access to videos of the agents' behaviors, indicating a gap in performance in today's vision foundation models, since humans rely only on videos to judge performance. 

In practice, Figure \ref{fig:vlm-prefs} shows how often the selected agents in the main experiment corresponded to the humans' top ranked agent within an iteration, the second best agent, the third best agent, and so on. Overall, the preferences from the agent selection step serve as an effective replacement for human expertise, promising an alternative to manual fitness metric design for automated reward function evaluation.

\subsection{The Trajectory Alignment Filter Relies on Data}
After collecting a dataset of reliable preferences, the TAC can be computed for a new reward function to measure its alignment with respect to the dataset and consequently the data generating process. Although it has not been theoretically proven, the intuition behind the trajectory alignment filter is that reward functions which are better aligned with the preference generator also produce agents that are ultimately preferred by the same generator. To verify this empirically, we reuse all preferences generated during the experiments in Section \ref{sec:main-exp} to randomly create varying sizes of preference datasets, denoted $\mathcal{D}^x$ where $x$ is the size of the dataset, all originating from the same preference generator, the VLM. Let $\mathcal{D}^\text{full}$ be the complete dataset of 400 preferences. For different values of $x$, we compute how accurate the TAC for a reward function is at predicting post-training preferences with the following equation:
\begin{align*}
    \text{Acc}(\mathcal{D}^x) &:= \frac{1}{|\mathcal{D}^\text{full}|}\sum_{(p_{ij}, \tau_i, \tau_j) \sim \mathcal{D}_{\text{full}}} p_{ij}\hat{p}_{ij} + (1 - p_{ij})(1-\hat{p}_{ij}) \enspace ,\\
    \hat{p}_{ij} &= 1 \enspace \text{if} \enspace \sigma(R_i, \mathcal{D}^x) > \sigma(R_j, \mathcal{D}^x) \enspace \text{else} \enspace 0 \enspace .
\end{align*}

\begin{table}[t]
    \centering
    \resizebox{\textwidth}{!}{
    \begin{tabular}{c|c|c|c|c}
        \hline
        &Human to humans & VLM to humans & VLM-$\tau$ to humans & VLM-$I$ to humans \\
        \hline
        \hline
        Agree rate &  78.54\%&  74.14\%  & 71.48\% & 62.96\%\\
        \hline
    \end{tabular}}
    \vspace{0.5em}
    \caption{\small Agreement rate of different preference sources with human experts. VLM, VLM-$\tau$, VLM-$I$ indicate different input modalities. Specifically, using both trajectory data and video data, trajectories only, and videos only respectively.}
    \label{tab:vlm-prefs}
    \vspace{-2em}
\end{table}

The results of this experiment shown in Figure \ref{fig:other-exps}a) lead to a concerning conclusion: the trajectory alignment coefficient is only a reliable indicator for success after a certain dataset size is reached. In our case, the predictive accuracy of the TAC is only comfortably better than random guessing after collecting around 100 preferences. Unfortunately, the experiments in Section \ref{sec:main-exp} only collected 10 preferences per iteration, and reached a maximum of four iterations. Therefore, the trajectory alignment filter was unlikely to be very beneficial in our results. However, the TAC can still be a very cheap method to quickly evaluate new reward functions provided a large enough dataset of preferences is available, either by bootstrapping an initial set of preferences or by scaling $M$ and/or $K$ towards larger values, which we leave for future work.

\subsection{On the Novelty of Generated Reward Functions}
\label{sec:reward-cor}
\begin{figure*}[t]
    \centering
    
    \subfloat[TAC Accuracy.]{
    \includegraphics[width=0.31\linewidth]{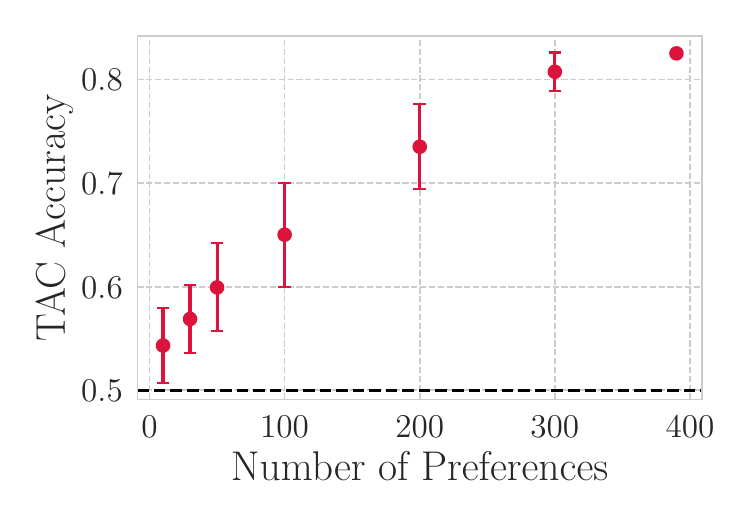}}
    \subfloat[Reward correlations.]{
            \includegraphics[width=0.33\linewidth]{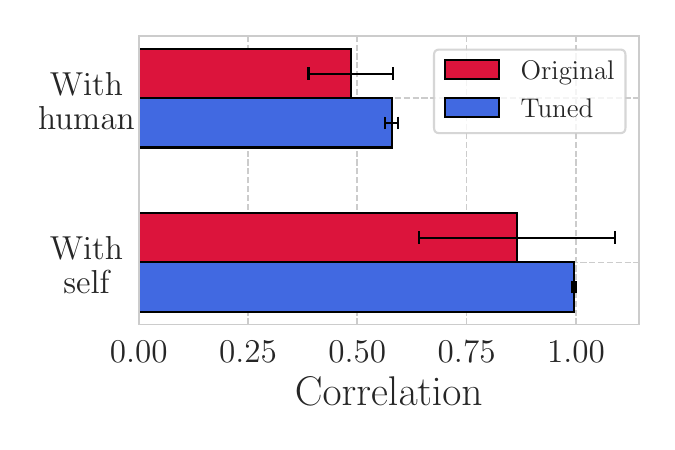}}
    \subfloat[Correlations vs final place.]
    {\includegraphics[width=0.33\linewidth]{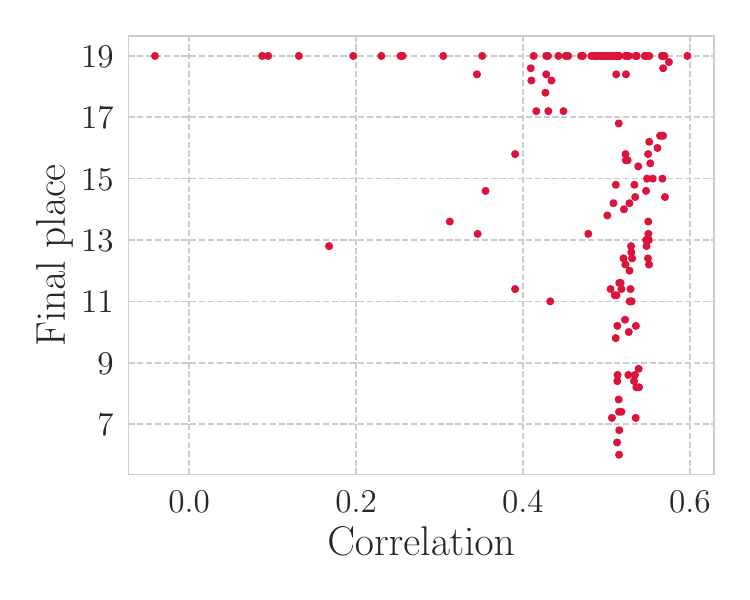}}
    \caption{\footnotesize \textbf{a)} The trajectory alignment coefficient's accuracy at predicting a reward function's post-training performance for varying sizes of preference datasets. The dashed line represents the baseline accuracy of a random guesser. \textbf{b)} The average correlation ($\pm \text{std}$) of all generated reward functions with respect to rewards from GT Sophy (with human) and the best LLM-based agent (with self) before (original) and after (tuned) optimizing the reward component weights towards the reference policy. \textbf{c)} The relationship between the correlation of a reward function with respect to the rewards from GT Sophy, and its intermediate policy's performance.}
    \label{fig:other-exps}
    \vspace{-1.5em}
\end{figure*}

To measure the \textit{novelty} of the generated rewards by our system, we compute the correlation of all the generated rewards from "\textbf{Ours}" against the rewards used to train GT Sophy. From a diverse set of trajectories (Appendix \ref{sec:cor-sample}), rewards are computed for \textit{all} LLM-generated reward functions and compared against a reference reward function to measure their correlations with the reference rewards. Two different reference reward functions are used to evaluate two different forms of novelty. First, the correlations are computed against the human designed rewards used to train GT Sophy (`With human') to see if  the generated rewards are novel with respect to humans. Second, the correlations are computed against a reward function generated by the LLM which achieved better performance than GT Sophy (`With self') to see if the generated rewards are  novel with respect to itself. The resulting correlations are shown in red in Figure \ref{fig:other-exps}b).

Recall that the generated reward functions are composed of multiple smaller components, each with a weight coefficient also generated by the LLM. To evaluate the qualitative novelty of reward functions, and not just potential differences in weights, we also compute the correlations with the \textit{tuned} reward functions with respect to the reference rewards, where the weight coefficients are tuned to maximize the correlation with the reference rewards (Appendix \ref{sec:tuning-weights}). With this change, the blue bars in Figure \ref{fig:other-exps}b) sees improved correlations, but the rewards are still relatively novel compared to the ones used to train GT Sophy. 

Further, notice that the average tuned correlation of the LLM's reward functions computed against itself is nearly 1.0, suggesting a very low degree of qualitative diversity within the generated  reward functions. In addition, since the  reference rewards in this case achieved better performance than the baseline GT Sophy agent, one potential solution to consistently generate high performing agents is to improve the weight tuning process. In our framework, no special care was taken towards tuning these weights: the LLM handled the weight selection concurrently with generating the reward function code. This remains an open problem to be solved in future research. 

Finally, the plot in Figure \ref{fig:other-exps}c) shows the relationship between a reward function's correlation with GT Sophy and the final placement achieved by its corresponding intermediate policy. The Pearson correlation between the $x$ and $y$ axis on this plot is $-0.177$; there is no strong relationship between rewards that are closer to the human designed ones and their performance, which further illustrates that rewards designed by humans are not necessarily optimal.

\subsection{Generating Novel Behaviors}
In our final set of experiments, we use our system to generate two novel behaviors not yet attempted with RL in GT 7. Using the same hyperparameters as before, the system is run on the following task descriptions: "\textit{race as fast as possible in reverse at all times while otherwise obeying standard motorsport rules}" and "\textit{race normally except the agent should drift as much as possible while otherwise obeying standard motorsport rules}". We found that a single seed was sufficient to achieve the desired behaviors in these cases based on human evaluations, although we cannot comment on their optimality. Videos of the trained policies for both novel behaviors can be found \href{https://sites.google.com/research.sony/autoreward-gt/home?authuser=2}{here}.

\section{Conclusion and Limitations}
Modern reinforcement learning algorithms present a powerful tool set to produce impressive agents through proper reward specification. In an environment like Gran Turismo, this led to the development of GT Sophy, a superhuman autonomous racing agent. While GT Sophy was trained without explicit human demonstrations, experts had to guide the agent's behavior through reward functions. These reward functions are an unnatural form of communication for most, especially those unfamiliar with RL. In this work, we demonstrated how today's foundation models can automate the reward design process for GT 7, creating a direct pipeline from text to policy by leveraging the large-scale domain expertise contained within a foundation model as a useful prior for learning. Our framework was shown to produce superhuman agents more consistently than prior work, as well as to generate novel reward functions and behaviors in a previously unexplored environment for automated reward design. Even though our system was only evaluated on the GT 7 environment, none of its components are specific to this environment. We leave the exploration of its broader applicability in other environments for future work. 

Our method shares several limitations with prior approaches to LLM-coding-based automated reward design. These approaches can be very computationally expensive to run, since the outer loop trains multiple RL agents in the inner loop and the foundation models themselves  must be quite large to write functional reward code. In addition, our system still relies on a human in the loop to achieve consistent results. Although we take steps to alleviate these limitations, for example with a secondary replay buffer and automatic VLM evaluations, further work can be done in this space towards cheaper and more reliable automated reward design.


{\small
\bibliography{references}

\begin{thebibliography}{10}

\bibitem{awais2025foundation}
M.~Awais, M.~Naseer, S.~Khan, R.~M. Anwer, H.~Cholakkal, M.~Shah, M.-H. Yang, and F.~S. Khan.
\newblock Foundation models defining a new era in vision: a survey and outlook.
\newblock {\em IEEE Transactions on Pattern Analysis and Machine Intelligence}, 2025.

\bibitem{badia2020agent57outperformingatarihuman}
A.~P. Badia, B.~Piot, S.~Kapturowski, P.~Sprechmann, A.~Vitvitskyi, D.~Guo, and C.~Blundell.
\newblock Agent57: Outperforming the atari human benchmark, 2020.

\bibitem{booth2023perils}
S.~Booth, W.~B. Knox, J.~Shah, S.~Niekum, P.~Stone, and A.~Allievi.
\newblock The perils of trial-and-error reward design: misdesign through overfitting and invalid task specifications.
\newblock In {\em Proceedings of the AAAI Conference on Artificial Intelligence}, volume~37, pages 5920--5929, 2023.

\bibitem{bradley1952rank}
R.~A. Bradley and M.~E. Terry.
\newblock Rank analysis of incomplete block designs: I. the method of paired comparisons.
\newblock {\em Biometrika}, 39(3/4):324--345, 1952.

\bibitem{carta2022eager}
T.~Carta, P.-Y. Oudeyer, O.~Sigaud, and S.~Lamprier.
\newblock Eager: Asking and answering questions for automatic reward shaping in language-guided rl.
\newblock {\em Advances in neural information processing systems}, 35:12478--12490, 2022.

\bibitem{chen2024end}
L.~Chen, P.~Wu, K.~Chitta, B.~Jaeger, A.~Geiger, and H.~Li.
\newblock End-to-end autonomous driving: Challenges and frontiers.
\newblock {\em IEEE Transactions on Pattern Analysis and Machine Intelligence}, 2024.

\bibitem{d2022sample}
P.~D'Oro, M.~Schwarzer, E.~Nikishin, P.-L. Bacon, M.~G. Bellemare, and A.~Courville.
\newblock Sample-efficient reinforcement learning by breaking the replay ratio barrier.
\newblock In {\em Deep Reinforcement Learning Workshop NeurIPS 2022}, 2022.

\bibitem{escontrela2023video}
A.~Escontrela, A.~Adeniji, W.~Yan, A.~Jain, X.~B. Peng, K.~Goldberg, Y.~Lee, D.~Hafner, and P.~Abbeel.
\newblock Video prediction models as rewards for reinforcement learning.
\newblock {\em Advances in Neural Information Processing Systems}, 36:68760--68783, 2023.

\bibitem{fuchs2021super}
F.~Fuchs, Y.~Song, E.~Kaufmann, D.~Scaramuzza, and P.~D{\"u}rr.
\newblock Super-human performance in gran turismo sport using deep reinforcement learning.
\newblock {\em IEEE Robotics and Automation Letters}, 6(3):4257--4264, 2021.

\bibitem{haarnoja2018soft}
T.~Haarnoja, A.~Zhou, P.~Abbeel, and S.~Levine.
\newblock Soft actor-critic: Off-policy maximum entropy deep reinforcement learning with a stochastic actor.
\newblock In {\em International conference on machine learning}, pages 1861--1870. Pmlr, 2018.

\bibitem{han2024autoreward}
X.~Han, Q.~Yang, X.~Chen, Z.~Cai, X.~Chu, and M.~Zhu.
\newblock Autoreward: Closed-loop reward design with large language models for autonomous driving.
\newblock {\em IEEE Transactions on Intelligent Vehicles}, 2024.

\bibitem{hazra2024revolve}
R.~Hazra, A.~Sygkounas, A.~Persson, A.~Loutfi, and P.~Z.~D. Martires.
\newblock Revolve: Reward evolution with large language models using human feedback.
\newblock {\em arXiv preprint arXiv:2406.01309}, 2024.

\bibitem{huang2024vlm}
Z.~Huang, Z.~Sheng, Y.~Qu, J.~You, and S.~Chen.
\newblock Vlm-rl: A unified vision language models and reinforcement learning framework for safe autonomous driving.
\newblock {\em arXiv preprint arXiv:2412.15544}, 2024.

\bibitem{kendall1945treatment}
M.~G. Kendall.
\newblock The treatment of ties in ranking problems.
\newblock {\em Biometrika}, 33(3):239--251, 1945.

\bibitem{klissarov2023motif}
M.~Klissarov, P.~D'Oro, S.~Sodhani, R.~Raileanu, P.-L. Bacon, P.~Vincent, A.~Zhang, and M.~Henaff.
\newblock Motif: Intrinsic motivation from artificial intelligence feedback.
\newblock In {\em NeurIPS 2023 Foundation Models for Decision Making Workshop}, 2023.

\bibitem{klissarov2024maestromotif}
M.~Klissarov, M.~Henaff, R.~Raileanu, S.~Sodhani, P.~Vincent, A.~Zhang, P.-L. Bacon, D.~Precup, M.~C. Machado, and P.~D'Oro.
\newblock Maestromotif: Skill design from artificial intelligence feedback.
\newblock {\em arXiv preprint arXiv:2412.08542}, 2024.

\bibitem{knox2023reward}
W.~B. Knox, A.~Allievi, H.~Banzhaf, F.~Schmitt, and P.~Stone.
\newblock Reward (mis) design for autonomous driving.
\newblock {\em Artificial Intelligence}, 316:103829, 2023.

\bibitem{kwon2023reward}
M.~Kwon, S.~M. Xie, K.~Bullard, and D.~Sadigh.
\newblock Reward design with language models.
\newblock {\em arXiv preprint arXiv:2303.00001}, 2023.

\bibitem{le2022coderl}
H.~Le, Y.~Wang, A.~D. Gotmare, S.~Savarese, and S.~C.~H. Hoi.
\newblock Coderl: Mastering code generation through pretrained models and deep reinforcement learning.
\newblock {\em Advances in Neural Information Processing Systems}, 35:21314--21328, 2022.

\bibitem{lee2025champion}
H.~Lee, T.~Seno, J.~J. Tai, K.~Subramanian, K.~Kawamoto, P.~Stone, and P.~R. Wurman.
\newblock A champion-level vision-based reinforcement learning agent for competitive racing in gran turismo 7.
\newblock {\em IEEE Robotics and Automation Letters}, 2025.

\bibitem{li2024auto}
H.~Li, X.~Yang, Z.~Wang, X.~Zhu, J.~Zhou, Y.~Qiao, X.~Wang, H.~Li, L.~Lu, and J.~Dai.
\newblock Auto mc-reward: Automated dense reward design with large language models for minecraft.
\newblock In {\em Proceedings of the IEEE/CVF Conference on Computer Vision and Pattern Recognition}, pages 16426--16435, 2024.

\bibitem{maeureka}
Y.~J. Ma, W.~Liang, G.~Wang, D.-A. Huang, O.~Bastani, D.~Jayaraman, Y.~Zhu, L.~Fan, and A.~Anandkumar.
\newblock Eureka: Human-level reward design via coding large language models.
\newblock In {\em The Twelfth International Conference on Learning Representations}, 2024.

\bibitem{macglashan2022sportsmanship}
J.~MacGlashan, A.~Devlic, and P.~MacAlpine.
\newblock Don't cross that line! how our ai agent learned sportsmanship.
\newblock \url{https://ai.sony/blog/Dont-Cross-That-Line!-How-Our-AI-Agent-Learned-Sportsmanship/}, 2022.
\newblock Accessed: 2025-08-21.

\bibitem{muslimani2025towards}
C.~Muslimani, K.~Johnstonbaugh, S.~Chandramouli, S.~Booth, W.~B. Knox, and M.~E. Taylor.
\newblock Towards improving reward design in rl: A reward alignment metric for rl practitioners.
\newblock {\em arXiv preprint arXiv:2503.05996}, 2025.

\bibitem{ng1999policy}
A.~Y. Ng, D.~Harada, and S.~Russell.
\newblock Policy invariance under reward transformations: Theory and application to reward shaping.
\newblock In {\em Icml}, volume~99, pages 278--287. Citeseer, 1999.

\bibitem{nikishin2022primacy}
E.~Nikishin, M.~Schwarzer, P.~D’Oro, P.-L. Bacon, and A.~Courville.
\newblock The primacy bias in deep reinforcement learning.
\newblock In {\em International conference on machine learning}, pages 16828--16847. PMLR, 2022.

\bibitem{olausson2023demystifying}
T.~X. Olausson, J.~P. Inala, C.~Wang, J.~Gao, and A.~Solar-Lezama.
\newblock Demystifying gpt self-repair for code generation.
\newblock {\em CoRR}, 2023.

\bibitem{ouyang2022training}
L.~Ouyang, J.~Wu, X.~Jiang, D.~Almeida, C.~Wainwright, P.~Mishkin, C.~Zhang, S.~Agarwal, K.~Slama, A.~Ray, et~al.
\newblock Training language models to follow instructions with human feedback.
\newblock {\em Advances in neural information processing systems}, 35:27730--27744, 2022.

\bibitem{remonda2021formula}
A.~Remonda, S.~Krebs, E.~Veas, G.~Luzhnica, and R.~Kern.
\newblock Formula rl: Deep reinforcement learning for autonomous racing using telemetry data.
\newblock {\em arXiv preprint arXiv:2104.11106}, 2021.

\bibitem{rocamonde2023vision}
J.~Rocamonde, V.~Montesinos, E.~Nava, E.~Perez, and D.~Lindner.
\newblock Vision-language models are zero-shot reward models for reinforcement learning.
\newblock {\em arXiv preprint arXiv:2310.12921}, 2023.

\bibitem{roziere2023code}
B.~Roziere, J.~Gehring, F.~Gloeckle, S.~Sootla, I.~Gat, X.~E. Tan, Y.~Adi, J.~Liu, R.~Sauvestre, T.~Remez, et~al.
\newblock Code llama: Open foundation models for code.
\newblock {\em arXiv preprint arXiv:2308.12950}, 2023.

\bibitem{singh2022progprompt}
I.~Singh, V.~Blukis, A.~Mousavian, A.~Goyal, D.~Xu, J.~Tremblay, D.~Fox, J.~Thomason, and A.~Garg.
\newblock Progprompt: Generating situated robot task plans using large language models.
\newblock {\em arXiv preprint arXiv:2209.11302}, 2022.

\bibitem{singh2009rewards}
S.~Singh, R.~L. Lewis, and A.~G. Barto.
\newblock Where do rewards come from.
\newblock In {\em Proceedings of the annual conference of the cognitive science society}, pages 2601--2606. Cognitive Science Society, 2009.

\bibitem{song2021autonomous}
Y.~Song, H.~Lin, E.~Kaufmann, P.~D{\"u}rr, and D.~Scaramuzza.
\newblock Autonomous overtaking in gran turismo sport using curriculum reinforcement learning.
\newblock In {\em 2021 IEEE international conference on robotics and automation (ICRA)}, pages 9403--9409. IEEE, 2021.

\bibitem{sontakke2023roboclip}
S.~Sontakke, J.~Zhang, S.~Arnold, K.~Pertsch, E.~B{\i}y{\i}k, D.~Sadigh, C.~Finn, and L.~Itti.
\newblock Roboclip: One demonstration is enough to learn robot policies.
\newblock {\em Advances in Neural Information Processing Systems}, 36:55681--55693, 2023.

\bibitem{sun2024large}
S.~Sun, R.~Liu, J.~Lyu, J.~Yang, L.~Zhang, and X.~Li.
\newblock A large language model-driven reward design framework via dynamic feedback for reinforcement learning.
\newblock {\em CoRR}, 2024.

\bibitem{tang2024rlrobots}
C.~Tang, B.~Abbatematteo, J.~Hu, R.~Chandra, R.~Martin-martin, and P.~Stone.
\newblock Deep reinforcement learning for robotics: A survey of real-world successes.
\newblock {\em arXiv preprint arXiv:2408.03539}, 2024.

\bibitem{tirumalareplay}
D.~Tirumala, T.~Lampe, J.~E. Chen, T.~Haarnoja, S.~Huang, G.~Lever, B.~Moran, T.~Hertweck, L.~Hasenclever, M.~Riedmiller, et~al.
\newblock Replay across experiments: A natural extension of off-policy rl.
\newblock In {\em The Twelfth International Conference on Learning Representations}, 2023.

\bibitem{wei2022chain}
J.~Wei, X.~Wang, D.~Schuurmans, M.~Bosma, F.~Xia, E.~Chi, Q.~V. Le, D.~Zhou, et~al.
\newblock Chain-of-thought prompting elicits reasoning in large language models.
\newblock {\em Advances in neural information processing systems}, 35:24824--24837, 2022.

\bibitem{wirth2017survey}
C.~Wirth, R.~Akrour, G.~Neumann, and J.~F{\"u}rnkranz.
\newblock A survey of preference-based reinforcement learning methods.
\newblock {\em Journal of Machine Learning Research}, 18(136):1--46, 2017.

\bibitem{wurman2022outracing}
P.~R. Wurman, S.~Barrett, K.~Kawamoto, J.~MacGlashan, K.~Subramanian, T.~J. Walsh, R.~Capobianco, A.~Devlic, F.~Eckert, F.~Fuchs, et~al.
\newblock Outracing champion gran turismo drivers with deep reinforcement learning.
\newblock {\em Nature}, 602(7896):223--228, 2022.

\bibitem{xietext2reward}
T.~Xie, S.~Zhao, C.~H. Wu, Y.~Liu, Q.~Luo, V.~Zhong, Y.~Yang, and T.~Yu.
\newblock Text2reward: Reward shaping with language models for reinforcement learning.
\newblock In {\em The Twelfth International Conference on Learning Representations}, 2024.

\bibitem{yang2024adapt2reward}
Y.~Yang, M.~Chen, Q.~Qiu, J.~Wu, W.~Wang, B.~Lin, Z.~Guan, and X.~He.
\newblock Adapt2reward: Adapting video-language models to generalizable robotic rewards via failure prompts.
\newblock In {\em European Conference on Computer Vision}, pages 163--180. Springer, 2024.

\bibitem{yu2023language}
W.~Yu, N.~Gileadi, C.~Fu, S.~Kirmani, K.-H. Lee, M.~G. Arenas, H.-T.~L. Chiang, T.~Erez, L.~Hasenclever, J.~Humplik, et~al.
\newblock Language to rewards for robotic skill synthesis.
\newblock In {\em Conference on Robot Learning}, pages 374--404. PMLR, 2023.

\bibitem{zhang2024orso}
C.~B.~C. Zhang, Z.-W. Hong, A.~Pacchiano, and P.~Agrawal.
\newblock Orso: Accelerating reward design via online reward selection and policy optimization.
\newblock In {\em ICML 2024 Workshop: Aligning Reinforcement Learning Experimentalists and Theorists}, 2024.

\end{thebibliography}
\bibliographystyle{abbrv}
}

\newpage
\appendix

\section{Ablation on Secondary Replay Buffer}
\label{sec:buffer-ablation}
\begin{figure}[!hbt]
    \centering
    \includegraphics[width=0.8\linewidth]{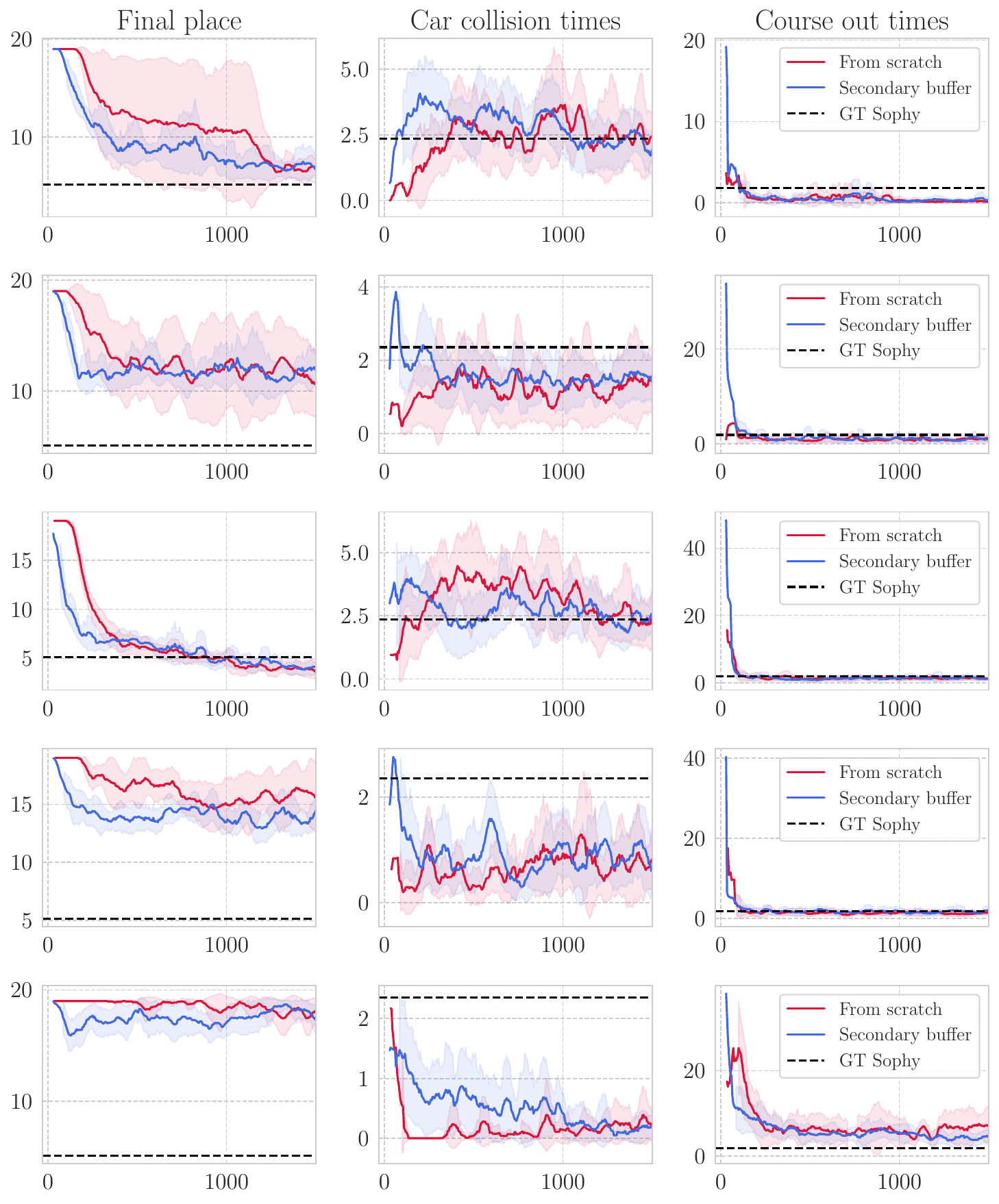}
    \caption{\footnotesize Training curves for five different generated final reward functions from \textbf{Ours} (rows). Each column shows a different training metric. Each reward function is trained with identical hyperparameters, except one is trained with a secondary replay buffer from the previous iteration, and one is trained from scratch. The $x$-axis for all plots are the number of epochs.}
    \label{fig:buff-abl}
\end{figure}

We conduct an additional ablation to see the impact of the secondary replay buffer on training. On five randomly chosen final agents from \textbf{Ours}, we train the agent for the full 1500 model clocks with and without a secondary seed buffer in Figure \ref{fig:buff-abl}. We observe that while the secondary replay buffer can be quite beneficial for earlier epochs, the policies ultimately converge to a similar behavior in either case.

\section{Reward Generation Prompts}
\label{sec:example}
We separate the task of generating rewards into two prompts. First, we ask the LLM to generate an english-readable overview of the rewards. Then, the answer is provided to the LLM again to generate the corresponding Python code. There is also a separate set of prompts for generating the initial reward functions, and generating subsequent reward functions which improve upon the previous iteration.

First the prompt used to generate the english-readable overview of the initial reward functions for the first iteration.
\begin{tcolorbox}[title=Initial English Rewards Generation, colback=gray!10, colframe=black!50]
\begin{lstlisting}
You are an expert in motorsports, Gran Turismo 7, and reinforcement learning.
I would like to design a reward function to train an agent to race in Gran Turismo 7.
The goal of the agent is to {goal}.

Help me design a reward function for the above behavior by separating the total reward into several component rewards to facilitate future evaluations and improvements. Each individual reward function component should generate a separate reward for each of the N cars on the track. Each reward function component should have the following function signature, and output a numpy array of shape (N, 1):

```def reward_component(current_obs: GTObs, prev_obs: GTObs, course: Course) -> np.ndarray:```

Describe in words the high level function of each different reward component. Do not provide any code.

The code describing the attributes of the GTObs and Course objects in the reward input is provided below. You are not obligated to use all attributes.

```\n{obs_code}\n```

Directly answer in the following format for each reward function component. The reward name should follow python function definition naming conventions.
<|reward name|> Name here <|reward name|>
<|reward description|> Description here <|reward description|>

Additional information:
- Do not provide any code, only provide the reward names and descriptions.
- The game designers have defined being out of bounds as having any of the four tires off the racing area.
- Keep the complexity and number of components to a minimum. It will be possible to further iterate and improve on the rewards at a later time.

\end{lstlisting}
\end{tcolorbox}

Then, the output from the previous prompt is given to the following prompt to generate the actual Python code used for RL training.

\begin{tcolorbox}[title=Initial Code Rewards Generation, colback=gray!10, colframe=black!50]
\begin{lstlisting}
You are an expert in motorsports, Gran Turismo 7, reinforcement learning, and python code generation.

I would like to implement the following reward components. Each component has a function name wrapped by <|reward name|>, and a description of the reward function wrapped by <|reward description|>.

{reward_components_desc}

Each individual reward function component should generate a separate reward for each of the N cars on the track. Each reward function component should have the following function signature, and output a numpy array of shape (N, 1) with a float data type:
```def reward_name(current_obs: GTObs, prev_obs: GTObs, course: Course) -> np.ndarray:```

Generate code for each of the described reward function components. The final reward will be a weighted sum of each of these component reward functions. Within the block of code, after all function definitions, suggest a weight coefficient for each component in the following format:
```weights = {{"first_reward_name": weight, "second_reward_name": weight, ...}```

The code describing the attributes of the GTOBs and Course objects in the reward input is provided below.
```\n{obs_code}\n```

Additional information:
- All reward function components should be contained within a single code block
- There should be no helper functions. The only function definitions within the generated block should be the reward components themselves and the weights dictionary. There should be no other function definitions.
\end{lstlisting}
\end{tcolorbox}

For all subsequent iterations, the following pair of prompts are used in a similar fashion to generate an english overview of the improvements to implement, and to generate the Python code once more. 

\begin{tcolorbox}[title=Subsequent Improvements Suggestions, colback=gray!10, colframe=black!50]
\begin{lstlisting}
You are an expert in motorsports, Gran Turismo 7, and reinforcement learning. We would like to train a reinforcement learning agent in Gran Turismo 7. The goal of the agent is to {goal}
Read the following code describing the attributes of the GTObs and Course objects, which are used to design reward functions.

```python
{obs_code}
```

To achieve the above goal, we design reward functions iteratively by improving on the previous reward functions. After every iteration, the reward functions are used to train an agent, and improved following a set of instructions based on feedback from the trained agent. Thus far, we have designed {num_iters} round(s) of reward function components based on the following instructions.

{all_english_rewards}

Finally, the most recent set of instructions were used to write the following most recent reward function components.

```python
{reward_code}
```

These reward components were used to train a reinforcement learning agent to {goal}. The total reward used to train the agent is a weighted sum of these rewards using the weights dictionary at the bottom of the code block.

We evaluated our agent every 100 epochs during training. Evaluations are done by placing the agent in a race against 19 other cars. The other cars are controlled by a built-in AI provided by the video game. The agent always begins the race in last place. An evaluation episode terminates after completing {max_laps} laps, or after {max_episode_duration} seconds has elapsed. The evaluation course has a maximum fixed_course_v of {course_max_v} meters. We track the total reward for each individual reward component over an evaluation episode. The rewards reported are already multiplied by their corresponding reward weights. The total values for each reward component for one evaluation episode is shown below as a list. They are recorded at intervals of 100 training epochs, showing the agent's progress during training for each reward component. Additionally, the return is also shown, which is simply the sum of all weighted reward components.

```
{moi_data}
```

After examining videos of the trained agent, we also provide the following *ground truth* feedback:
{human_feedback}

Carefully analyze the current reward function code in conjunction with the training progress for each reward component. First, provide a short summary of the training run given the available information. This summary should be wrapped by a <|training summary|> tag.
Then, take everything into consideration to suggest improvements to the current reward functions. You may do one of four things, which should follow the format:
1. Modify reward weights: <|modify reward weight|> existing reward function name <|modify reward weight|>\n <|description|> {{Current weight}} --> {{New weight}}, followed by a short explanation. <|description|>
2. Modify the implementation of existing reward functions: <|modify reward|> existing reward function name <|modify reward|>\n<|description|> Description of modifications <|description|>
3. Completely remove existing reward functions: <|remove reward|>  existing reward function name <|remove reward|>
4. Add new reward function: <|new reward|> new reward function name <|new reward|> \n<|description|> Description here <|description|>

At the end, summarize all suggestions wrapped by a <|suggestions summary|> tag.

Additional information:
- Do not provide any code.
- The game designers have defined being out of bounds as having any of the four tires off the racing area.
- Limit the quantity and complexity of suggested changes, as performing too many significant changes at once can lead to unexpected results. For example, significantly modifying existing reward components or adding new reward components should be avoided unless absolutely necessary.
\end{lstlisting}
\end{tcolorbox}

\begin{tcolorbox}[title=Subsequent Improvements Implementation, colback=gray!10, colframe=black!50]
\begin{lstlisting}
Generate code for each reward function component once more, this time incorporating all suggested changes. Remember to include the weights for the reward components in a dictionary at the end of the code block.
Additional information:
- All reward function components should be contained within a single code block
- There should be no helper functions. The only function definitions within the generated block should be the reward components themselves and the weights dictionary. There should be no other function definitions.
\end{lstlisting}
\end{tcolorbox}

\section{Eliciting Preferences from the VLM}
\label{sec:vlm-prompts}
Evaluation videos of the trained agents range anywhere from 10 to 30 minutes to complete four laps of the designated track. We only provide the first lap or 3 minutes (whichever comes first) of video data for the VLM to evaluate at a frame rate of 10 frames per second. In order to reduce the context length input to the VLM, we further split the video into 15 seconds clips. The VLM then provides a description of the two agents' behaviors in each clip, then all descriptions are fed into an LLM as text for the LLM to decide which agent was preferred. We found this to be the best compromise in order to analyze long-form videos at a relatively high frame rate, evaluating proper behavior in Gran Turismo 7 requires high frame rate fidelity. The prompt used by the VLM to generate text summaries of a pair of 15 seconds clips is given below.

\begin{tcolorbox}[title=Summarizing Videos Into Text, colback=gray!10, colframe=black!50]
\begin{lstlisting}
You are an expert in motorsports and Gran Turismo. We trained two agents to play the Gran Turismo video game. The primary goal of both agents is to {goal}.

We trained two agents to play the Gran Turismo video game. The primary goal of both agents is to {goal}. You will be given information about both agents' trajectories, by receiving frame-by-frame data coming directly from the environment's API. The data will be coming at {fps} frames per second, and represents a {clip_length} second video clip of each agent's behavior. Documentation about each frame's data is given here:\n {variables_documentation}\n

The first video clip is shown below. 
{frames_and_observations_1}

Now watch this second video clip
{frames_and_observations_2}

Describe and compare in detail any and all of the agents' behaviors in each clip as it pertains to the desired goal. 
\nAdditional information: \n
 - The game designers have defined being out of bounds as having any of the four tires off the racing area.\n
 - There should be enough detail to evaluate the agents without watching the video. \n
 - Remain as objective as possible for all descriptions, 
as to not bias anyone towards whether the goal was achieved or not for each agent.\n
 - Directly answer in the following format: description of agent 1 in clip 1, description of agent 2 in clip 2, and a comparison between the two clips.
    
\end{lstlisting}
\end{tcolorbox}

Then, the text descriptions of all 15 seconds clips are combined to create a long description of the entire 3 minute video. The prompt used to gather the preference from an LLM over the stitched text-based descriptions of the two agent's behaviors is shown:

\begin{tcolorbox}[title=Getting Preference over Agents, colback=gray!10, colframe=black!50]
\begin{lstlisting}
You are an expert in motorsports and Gran Turismo. We trained two agents to play the Gran Turismo video game. The primary goal of both agents is to {goal}.

The final behavior of the two agents were recorded into two {video_duration} second videos, recorded at {fps}. The two videos, one for each agent, were then analyzed and compared clip by clip for each agent. The first clip always describes the first agent, and the second clip always describes the second agent.
The result of comparing every pair of clips in the complete video is shown below.
\n\n```{full_comparison_text}```\n\n
    
After carefully analyzing the descriptions of each agent's behavior for each clip, declare which agent is overall preferable with respect to achieving or making progress towards the desired goal for the entire video. 
Answer in the following format: \n
'("preferred_agent": 1) if the first agent is preferred or\n'
'("preferred_agent": 2) if the second agent is preferred.\n'
Include also a brief explanation for your answer.\n
\end{lstlisting}
\end{tcolorbox}

The LLM was not given the option to provide ties. We found it was important to use a contrastive approach to gather clip descriptions. To do so, the VLM was always fed two clips side-by-side for more accurate descriptions of an agent's behavior, since inaccuracies are not as important as getting a relative sense of which agent was better.

\section{Replacing Human Feedback with an LLM's}
\label{sec:closed-loop-system}
In order to conduct the closed-loop experiments from the \textbf{No Feedback} set of runs, we replaced all human feedback in the prompts with one generated automatically by an LLM following the agent descriptions obtained in Appendix \ref{sec:vlm-prompts}. The prompt used to gather this description is given here.
\begin{tcolorbox}[title=Getting Agent Behavior Descriptions, colback=gray!10, colframe=black!50]
\begin{lstlisting}
Carefully analyze the descriptions of each agent's behavior for each clip. Then, provide a detailed summary of agent {agent_to_describe}'s behavior only, especially as it relates to the stated goal. 
Additional notes: \n
 - The game designers have defined being out of bounds as having any of the four tires off the racing area.\n
 - Remain objective for all descriptions."
 Do no give opinions on whether certain behaviors are 'good' or 'bad'.\n
 - Make no mention of the other agent in the summary.\n
 - Start your response with '## Agent Summary'. 
Simply refer to the agent as 'the agent', instead of 'agent 1' or 'agent 2'. 
\end{lstlisting}
\end{tcolorbox}

\begin{table}[hbt!]
\centering
\caption{Relevant Training Parameters}
\label{tab:training_params}
\begin{tabular}{lr}
\hline
\textbf{Hyperparameter} & \textbf{Value}\\
\hline
Optimizer & Adam \\
Policy learning rate & 1.25e-5\\
Critic learning rate & 2.5e-5\\
Entropy coefficient $\alpha$ & 0.01 (fixed) \\
Discount & 0.9896 \\
n-step & 7 \\
Number of Quantiles &  32 \\
Parameter copy $\tau$ & 0.005 \\
Secondary replay buffer sampling ratio & 0.2\\
Secondary replay buffer size &  100 000\\
Replay buffer size & 1 000 000 \\

\end{tabular}
\end{table}

\section{Reinforcement Learning Hyperparameters}
We present the relevant hyperparameters used to train all agents in Table \ref{tab:training_params}. Hyperparameters were selected from the baseline GT Sophy agent. Additional hyperparamters used only for our method (secondary replay buffer) were selected following a small grid search.

\section{GT Sophy Rewards}
\label{sec:gt-sophy-rewards}
The following list describes the reward components used by the baseline Sophy. In general, we found that the generated reward components captured the same categories of rewards and penalties: progress, collisions, track violations, and passing. Penalties for steering smoothness were generated occasionally.

\begin{itemize}
    \item \textbf{progress}: A reward for the amount of progress made along the track center-line between observations.
    \item \textbf{off\_course\_penalty}: Penalty for the maximum amount of time that any one tire was off course since the last observation.
    \item \textbf{wall\_penalty}: Penalty for hitting the wall.
    \item \textbf{rel\_v2\_collision\_penalty}: A complex penalty that tries to determine collision fault. If our car is at fault it is penalized by the square of the difference in velocity vectors.
    \item \textbf{constant\_car\_collision}: A penalty for any collision with other cars.
    \item \textbf{passing\_reward}: A reward for gaining position on opponents and a penalty for losing position on opponents. Only considers opponents within certain distance range.
    \item \textbf{steering\_change\_cost}: A penalty for the magnitude of change in steering angle. Used to discourage large sudden changes.
    \item \textbf{acted\_steering\_history\_cost}: A penalty for making many changes in steering direction in a short window of time.
\end{itemize}

\section{LLM Generated Reward Functions}

Instead of providing the raw reward functions generated by the LLM, we conduct an informal qualitative analysis of the generated functions to better understand the diversity of components, and how they relate to the human designed reward functions in the previous section. We identified the following ten most popular categories of reward components generated by the LLM. For each category, there can be several different implementations and approaches to penalizing or rewarding the agent.

\begin{itemize}
    \item \textbf{collision\_avoidance}: Reward to discourage collisions through instantaneous penalties or rewarding continuous safe driving.
    \item \textbf{track\_boundary}: Penalty for going out of bounds.
    \item \textbf{forward\_progress}: Reward for making forward progress through the track.
    \item \textbf{driving\_smoothness}: Penalty for inhuman-like movements, or jerky movements. 
    \item \textbf{clean\_overtake}: Reward for overtaking an opponent while avoiding any collisions.
    \item \textbf{rank\_improvement}: Reward for improving the agent's current position on the track relative to other opponents.
    \item \textbf{optimal\_path\_following}: Reward for staying near the centerline of the track.
    \item \textbf{finish\_line\_bonus}: Reward for crossing the finish line.
    \item \textbf{gap\_closure}: Reward for closing in on the next closest opponent ahead of the agent.
    \item \textbf{off\_track\_overtake}: Penalty for overtaking opponents while outside the track boundaries
\end{itemize}

\begin{table}[ht]
\centering

\label{tab:reward_percent}
\begin{tabular}{l c}
\toprule
Category & Percent of Runs \\
\midrule
collision\_avoidance & 100.0\% \\
track\_boundary & 100.0\% \\
forward\_progress& 100.0\% \\
driving\_smoothness & 91.0\% \\
clean\_overtake& 59.0\% \\
rank\_improvement & 58.5\%\\
optimal\_path\_following & 55.0\%\\
finish\_line\_bonus & 15.5\%\\
gap\_closure & 8.5\%\\
off\_track\_overtake & 3.5\%\\
\bottomrule
\end{tabular}
\vspace{0.5em}
\caption{Different categories of reward components generated by the LLM from all seeds in \textbf{Ours}, and the percentage of generated reward functions containing some form of the specified component. In total, 200 reward functions were generated and analyzed, twenty for each of the ten seeds run for our method. }
\end{table}



\section{Additional Notes on Reward Correlations and Performance}
\subsection{Sampling Trajectories for Correlation Experiments}
\label{sec:cor-sample}
Trajectories were generated using one of the 300 epochs runs generated by our system. For every 10th epoch we used the saved policy to run a one lap race against 19 built-in-ai agents, with our target policy starting in 10th position. This gives a range of experiences resulting from the varying level of skill produced at different stages of training. Further, by placing our agent in the middle of the track this encourages more interaction with other cars even when the racing skill might be low.

\subsection{On Tuning the Reward Weight Coefficients}
\label{sec:tuning-weights}

As mentioned in Section \ref{sec:reward-cor}, one potential improvement to the framework is to focus on tuning the weight coefficients of the reward components themselves instead of the qualitative nature of the rewards. This is due to the observation that nearly all reward functions generated by the LLM, even those with poor performance, can have their weights modified to achieve near perfect correlation with a reward function with \textit{good} performance. In this section, we test the hypothesis that tuning the reward weights alone can be sufficient to achieve more consistent results.

\paragraph{Experimental setup} First, we select the best performing seed from our method and use its reward function as a reference reward. Then, we tune the reward weights from the final reward function of two other random seeds (out of ten) with poor performance with respect to the reference rewards with good performance. The reward weights are tuned with stochastic gradient descent from the trajectories collected in Appendix \ref{sec:cor-sample} in two stages: the weights are initially optimized to maximize correlation, then the weights are multiplied by a constant that is optimized to minimize the mean squared error to account for differences in scale. Finally, the agent is re-trained with the tuned reward weights.

\begin{figure}[!hbt]
    \centering
    \includegraphics[width=0.99\linewidth]{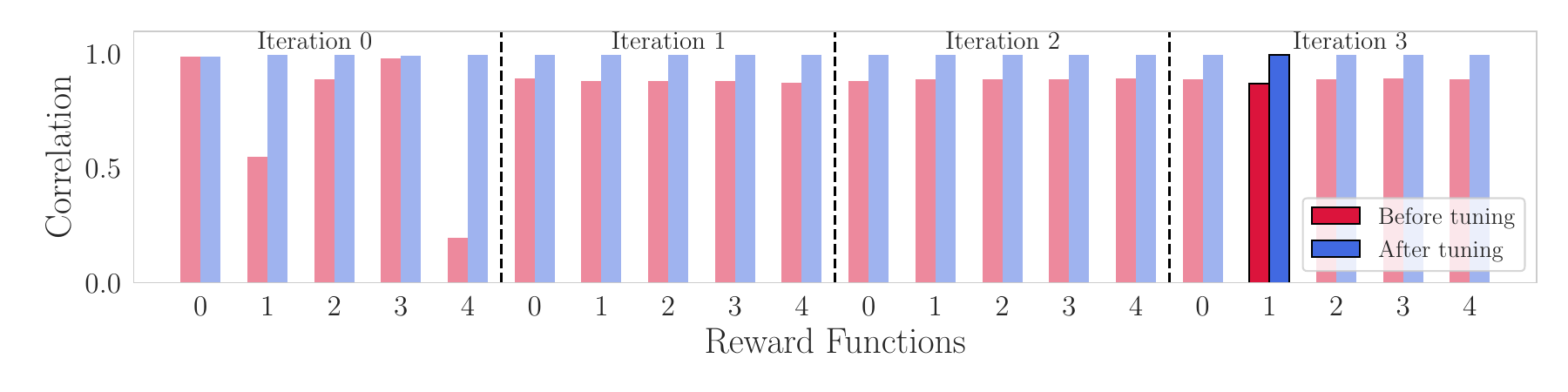}
    \includegraphics[width=0.99\linewidth]{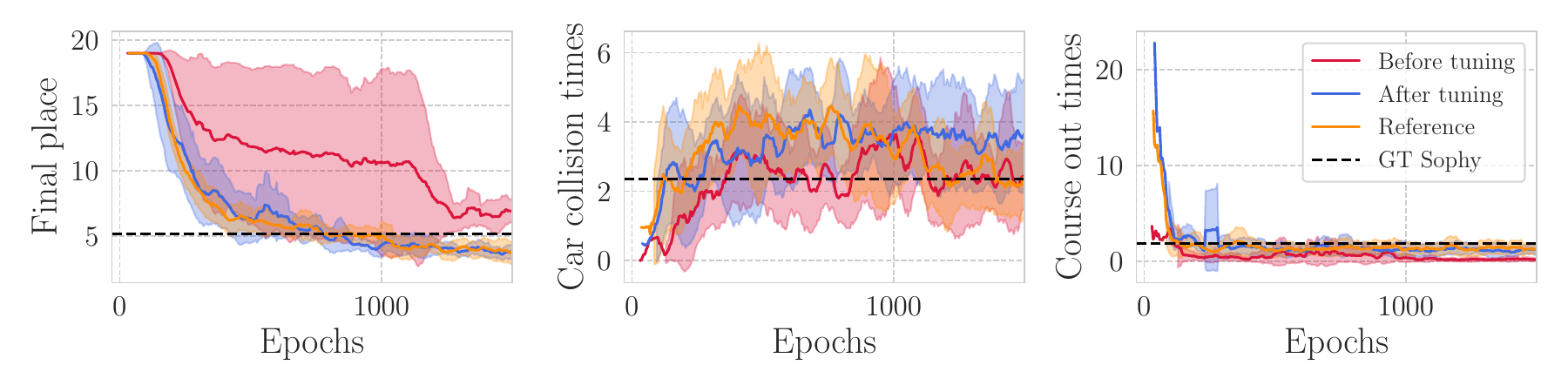}
    \caption{\footnotesize \textbf{Tuning weight coefficients can improve performance.} \textit{Top:} Correlations of generated reward functions for all iterations of a single seed  of \textbf{Ours} both before and after tuning the weights with respect to the best LLM-based agent's reward function. The darker bars represent the chosen reward function for the final iteration. \textit{Bottom:} Training curves of the final chosen reward function of the same seed. We report the mean final place, car collision times in seconds, and course out times in seconds across three seeds ($\pm$ std)}
    \label{fig:cor-perf-purple-dole}
\end{figure}

\begin{figure}[!hbt]
    \centering
    \includegraphics[width=0.99\linewidth]{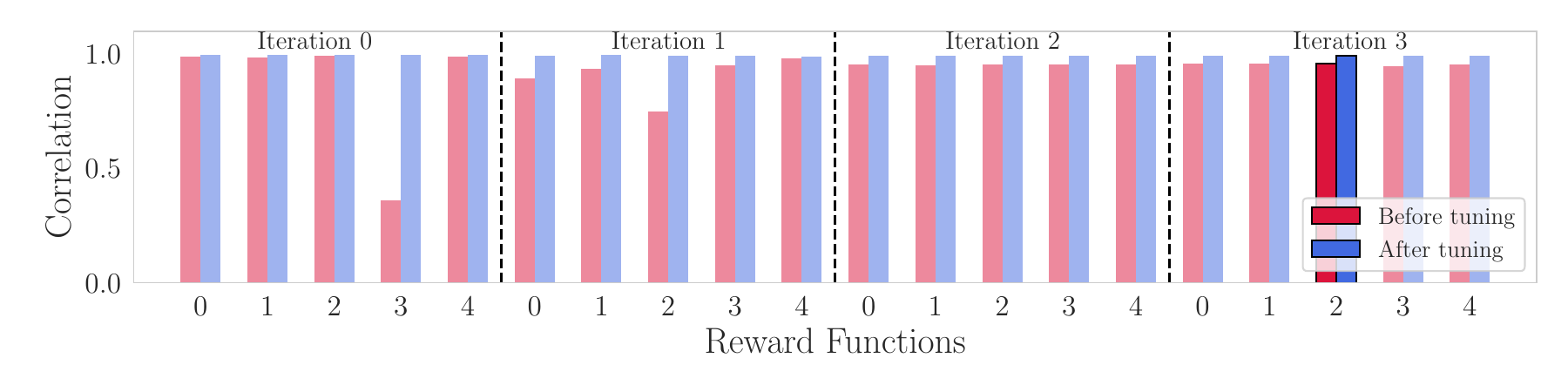}
    \includegraphics[width=0.99\linewidth]{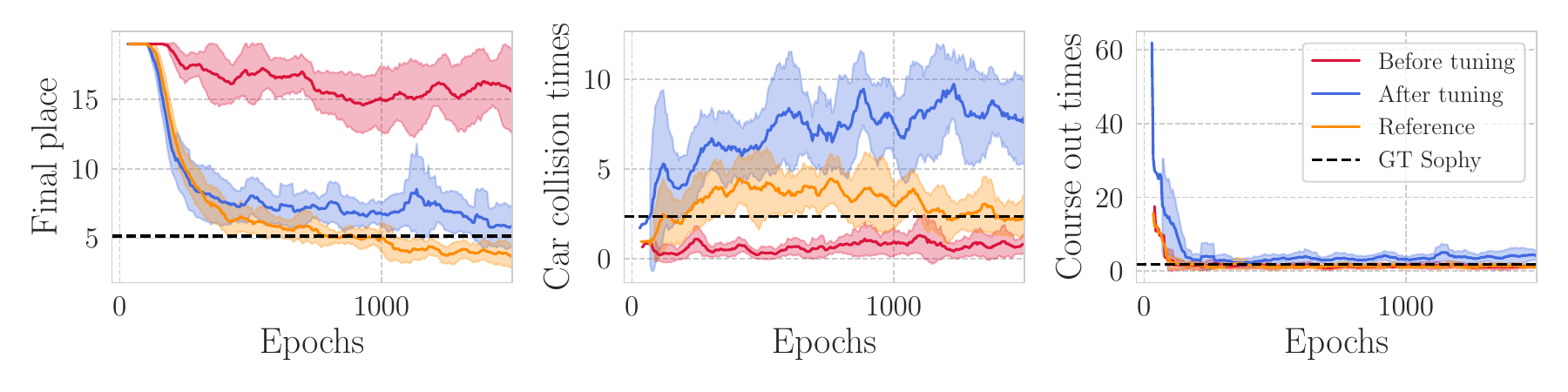}
    \caption{\footnotesize \textbf{Tuning weight coefficients is not always sufficient.} \textit{Top:} Correlations of generated reward functions for all iterations of a single seed  of \textbf{Ours} both before and after tuning the weights with respect to the best LLM-based agent's reward function. The darker bars represent the chosen reward function for the final iteration. \textit{Bottom:} Training curves of the final chosen reward function of the same seed. We report the mean final place, car collision times in seconds, and course out times in seconds across three seeds ($\pm$ std)}
    \label{fig:cor-perf-undue-suit}
\end{figure}

\paragraph{Results} Figures \ref{fig:cor-perf-purple-dole} and \ref{fig:cor-perf-undue-suit} report the results for two sub-optimal seeds. Although the tuned rewards are almost perfectly correlated with the reference reward, there may still be a gap in performance in the tuned rewards performance. For example, the results in Figure \ref{fig:cor-perf-undue-suit} show improved placements, but at the cost of increased collisions that do not quite match the expected performance from the reference reward function.

\end{document}